# Unsupervised Segmentation of Action Segments in Egocentric Videos using Gaze

Hipiny, H. Ujir, J.L. Minoi, S.F. Samson Juan and M.A. Khairuddin
*Faculty of Computer Science and Information Technology*
*Universiti Malaysia Sarawak*
*Kota Samarahan, Sarawak, Malaysia*

M.S. Sunar
*Faculty of Computing*
*Universiti Teknologi Malaysia*
*Johor Bahru, Johor, Malaysia*

***Abstract*—Unsupervised segmentation of action segments in egocentric videos is a desirable feature in tasks such as activity recognition and content-based video retrieval. Reducing the search space into a finite set of action segments facilitates a faster and less noisy matching. However, there exist a substantial gap in machine's understanding of natural temporal cuts during a continuous human activity. This work reports on a novel gaze-based approach for segmenting action segments in videos captured using an egocentric camera. Gaze is used to locate the region-of-interest inside a frame. By tracking two simple motion-based parameters inside successive regions-of-interest, we discover a finite set of temporal cuts. We present several results using combinations (of the two parameters) on a dataset, i.e., BRISGAZE-ACTIONS. The dataset contains egocentric videos depicting several daily-living activities. The quality of the temporal cuts is further improved by implementing two entropy measures.**

## I. INTRODUCTION

Unsupervised segmentation of *action segments* is a useful feature to have in activity recognition and content-based video retrieval. Both tasks involve the matching of discovered temporal segments inside a video to a learned action/scene. Thus, it is imperative for each discovered temporal segment to be quasi-identical with its corresponding learned action/scene, in term of the underlying structure of data.

Human activities naturally consist of non-mutually exclusive events. For example, a typical list of events when drinking coffee would consist of the following verb-object pairs: "Look-at cup", "Grab cup", "Raise cup", "Drink coffee", and "Place cup". These events, or *action segments*, may overlapped, abandoned prematurely, or paused for another action to occur, only to be resumed later. A person may continue drinking coffee whilst placing the cup onto a desk (overlapping events), initiating the motion of the active hand to grab the cup only to end it abruptly (abandoned prematurely), and pausing momentarily to answer a phone call (paused for another action, only to be resumed later). Nevertheless, the method presented in this work considers action segments as mutually exclusive events to simplify the problem.

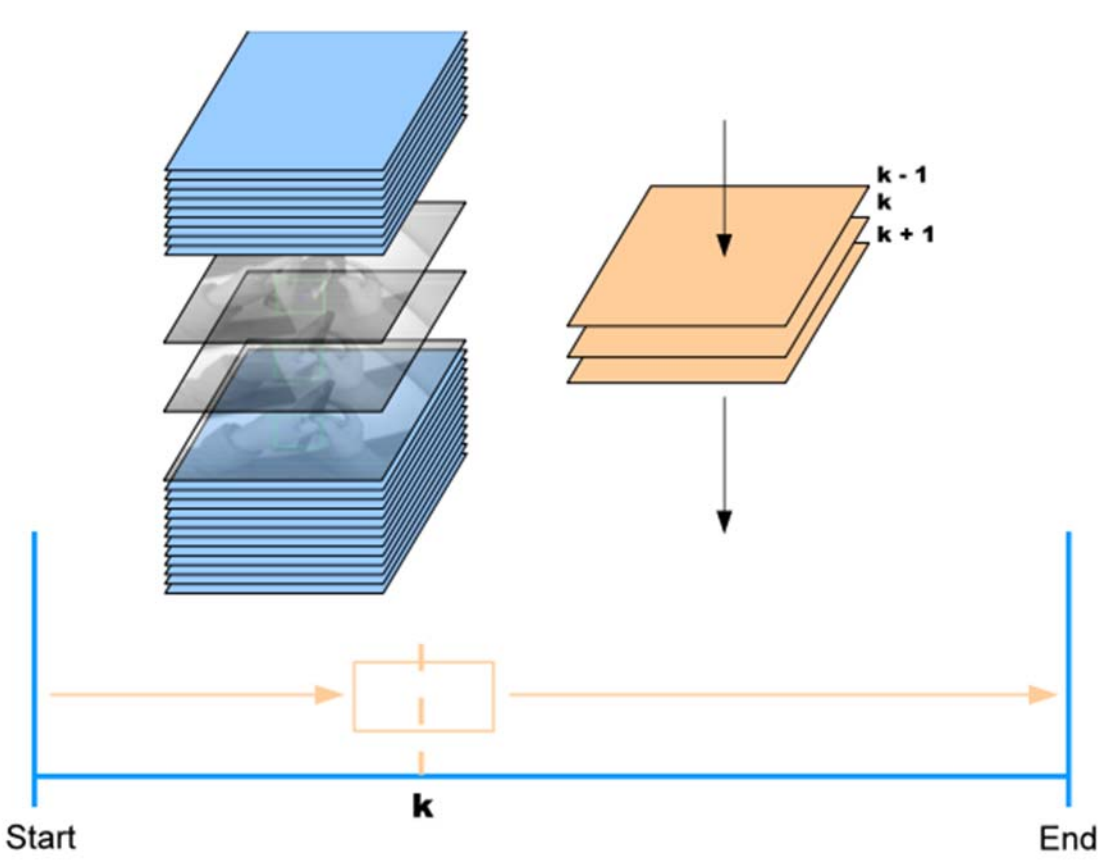

Fig. 1. A visual explanation of our temporal segmentation method, taken from [10]. Using a 9-frame sliding window (the above example uses n = 3 frames for visual clarity), we track two motion-based parameters, i.e., fpr and sdm, inside 9 successive frames (i.e., $I_{k-4}$, $I_{k-3}$, $I_{k-2}$, $I_{k-1}$, $I_k$, $I_{k+1}$, $I_{k+2}$, $I_{k+3}$, $I_{k+4}$). For each parameter, the 9 individual values are sorted and the median is chosen as the representative value for frame k.

Unsupervised segmentation and learning of motion primitives were addressed in both multimedia [25], [7], [15], and computer vision [34], [1], [12], [38] communities. A more recent work by Krüger et al. [19] considers action as a series of smaller, atomic motion primitives. These motion primitives are not required to have any semantic meaning, only for each to have a distinctive motion data. This finer representation allows overlapping events to be detected separately. However, results from [19] were obtained using non-egocentric datasets (stationary sensors). Our work is more similar to [16] and [32], as both evaluated their methods against egocentric datasets.

In supervised approaches, action segments are segmented a priori before further operations. Manual segmentation is performed with the understanding of natural segments inside an activity. In unsupervised approaches, temporal cuts are found by analysing cues inside the video without any prior knowledge of models. Thus, several assumptions must be made regarding the underlying structure of data at temporal cut points to differentiate them from the rest of the frames.

According to [2], [22], [13] and [9], gaze is closely related to a person's immediate cognitive goal. Ballard et al. [2] presented evidences of a tight-linkage in time between a person's gaze fixations and his or her actions. A person performing a task tend to acquire a specific visual information, just at the point it is required. These findings motivated us into using *Gaze region*, i.e., region-of interest containing a gaze fixation, to prime for clues for temporal cuts. We define a *Gaze region* as a $56 \times 56$ window centered around the gaze fixation point.

We observed that *Gaze regions* are usually registered when the egocentric camera is relatively stationary. A large camera motion induces saccades as the eyes move in a rapid fashion from a fixation point to another. Since no gaze fixation is registered during a saccade [6], a majority of the background features found inside a Gaze region would experience minimal motion (as induced by the camera). As for the foreground features (i.e., belonging to the active hand(s) and the currently-manipulated object), the translation, rotation and scaling would typically be intense. Based on this observation, we demonstrate the use of two simple parameters derived from motion features tracked across successive *Gaze regions*, i.e., Foreground Pixel Ratio ($fpr$), and Standard Deviation of Motions ($sdm$), for discovering temporal cuts. A visual explanation of our method is provided in Figure 1.

## II. PREVIOUS WORK

### *A. Autonomous Segmentation of Action Segments*

1) Supervised: Supervised segmentation of motion data can either be performed using known poses [4], [8], [26], known templates [27], [29], [21], [35] and motion synthesis [18], [17].

Beaudoin et al. [4] generates a motion-motif graph from a sequence of motion capture (Mo-cap) data by clustering similar motions together with their encompassing motion graph. Chew et al. [8] uses fuzzy clustering to compress motion data based on data similarity within the anatomy structure of a virtual character model and the temporal coherence within the motion data itself. López-Méndez et al. [26] generates pose clusters associated with unlabeled actions. Lv et al. [27] learns discriminative templates from manually-labelled segments using HMMs and AdaBoost. Müller et al. [29] introduces geometric-based templates to perform adaptive matching of unknown segments. In [21] and [35], segmentation using limited exemplars is made possible by learning the intrinsic regularities of the Mo-cap data. In [18], motion graphs are constructed from clips of Mo-cap data. In [17], motion elements are queried from large datasets, and then blended according to pre-defined user constraints.

2) Unsupervised: Earlier works on unsupervised segmentation of human motion data focused on videos captured using a stationary sensor. Barbicˇ et al. [3] introduces a PCA-based method to partition motion sequences into individual segments. By temporally-tracking the quality of local PCA models, new models are defined when the data variance in existing PCA models exceeds a threshold. In [14], a combination of PCA and linear regression is used to derive a metric for segmenting unknown motion sequences. Krüger et al. [19] segments motion data by aligning segregated feature trajectories, before clustering them into groups based on a similarity measure. The final grouping does not necessarily need to have any semantic meaning.

Similar to our work, Kitani et al. [16] and Poleg et al. [32] deal with the unsupervised segmentation of egocentric videos. Both are based on tracking discontinuities of optical flow-based features. Kitani et al. [16] proposes a novel method for the unsupervised learning of ego-actions. Successive video frames with identical camera motion are grouped together as a single ego-action, each represented as a simple histogram based on the optical flow direction/magnitude and frequency. The use of camera motion's discontinuities to determine a temporal cut is possible since their dataset specifically contains extreme egocentric sport videos, i.e., outdoor activities that involve sudden and large on-body camera motions. Our dataset contains daily-living activities, which involve a lot subtler camera motions. Poleg et al. [32] uses displacement vectors, similar to optical flows, for long-term temporal segmentation. Unlike ours, the dataset used in [32] contains activities of relatively simple ego-motion, such as walking and wheeling.

### *B. Gaze*

Multiple studies [23], [22], [24], [30], [9] from Vision research had presented strong evidences linking gaze and activity. Yarbus [37] made an observation that "seeing" is linked to the cognitive goals of the observer. In [24], a subject's gaze was recorded during several kitchen-based activities. It was reported that 80% of the subject's gaze fixations fell on task-relevant objects. Also, gaze fixations are found to be tightly-linked in time to the progress of a task. Ballard et al. [2] introduces the term "Just-in-Time" to refer to this very strategy of acquiring specific visual clue just at the moment it is required in the task. This characteristic was also observed by Turano et al. [36].

TABLE I. THE BRISGAZE-ACTIONS DATASET

| Activity | Learned actions |
| --- | --- |
| Prepare to watch a DVD | "grab-dvd-case", "open-DVD-case", "take-out-DVD", "insert-DVD-into-slot", "grab-headphones/plug", "plug-into-jack", and "press-key-combo". |
| Cook noodles | "grab-pan", "fill-pan-with-tap-water", "place-pan-on-hob", "put-in-noodle", "switch-on/off-hob", and "pour-cooked-noodles-into-bowl". |
| Prepare a cup of tea | "fill-kettle", "place-kettle", "switch-on/off-hob", "grab-mug", "open-tea-pack", "take-out-and-put-tea-satchet-into-mug", "grab/open-sugar-jar", "scoop-sugar", "pour-kettle", and "stir". |
| Prepare a bottle of milk | "open-milk-bottle", "grab/open-flask", "pour-flask", "grab/open-milk-pack", "scoop-milk-powder", "screw-on-teat", and "shake-milk-bottle". |

## III. DATASET

In our experiments, we used a dataset of egocentric videos, i.e., BRISGAZE-ACTIONS, first introduced in [10]. The dataset is actually a subset of the larger BRIS-GAZE dataset [11], containing a reduced number of activities. The dataset contains 72 egocentric videos of 4 daily-living activities. Each activity was performed by 6 adult participants in 3 repetitions, see Table I.

These activities were chosen because all are highly-structured with weak constraints on actions. For example, action "Stir teaspoon" can only realistically be performed in a couple of ways; all requiring the mug to be vertical and the active hand holding the teaspoon, stirring it whilst it is positioned inside the mug. These weak constraints increase the likelihood of a successful matching between action segments and the learned actions.

To provide the ground truths, we manually segmented all learned actions inside each egocentric video. Based on a constant definition of the start and end of each learned action, we labelled the respective frames. The ground truths are used to evaluate our unsupervised segmentation results.

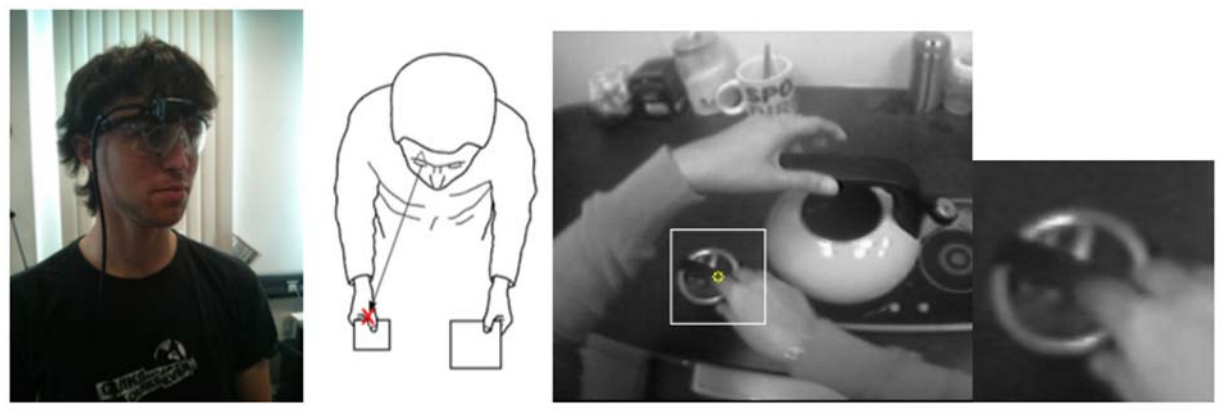

Fig. 2. The setup. All images sourced from [10].

Figure 2 shows the user setup during our experiments. Participants wear an ASL Mobile Eye device [20].

## IV. ENTROPY MEASURES

We employ two entropy measures to filter candidate frames. The goal is to improve the quality of temporal cuts, making sure that they closely resemble the natural temporal cuts of learned actions.

### A. Edge Ratio

Due to the arbitrary nature of gaze, some of the registered *Gaze regions* turned out to be flat-textured. Occasionally, participants glanced at the walls or on the desk surface for a period of time. These fixations do not serve any cognitive purposes therefore are not relevant to our task. We eliminate these flat-textured *Gaze regions* by first obtaining the edge image using Canny edge detector [5].

A threshold value of the edge ratio, $\varphi$,

$$\varphi = \frac{nzp}{56\times 56} \tag{1}$$

is used to reject *Gaze regions* with minimal number of edges, where the non-zero pixels, $nzp$, refers to the edge pixels' count (post-Canny). The *gaze region* has a width and height of 56 pixels.

### B. Color Histogram-based Hand Detection

To further increase the quality of temporal cuts, we only consider a candidate frame if the *Gaze region* contains human hand(s). Since actions inside the dataset are defined as verb-object pairs, we assume each action must start and end with a frame containing human hand(s). The verb-object pairing implies that an action involves the manipulation of an object by the participant's hand(s). This requirement is not enforced on the in-between frames due to the arbitrary nature of gaze.

We employ a color histogram-based approach, as used in [28], to detect the presence of human hand(s) inside a *Gaze region*. This method suffices as we only need a crude estimate of the skin area.

We populate a two-dimensional color histogram. A UV color space is used to provide invariance to illumination changes by manually sampling the UV values of each participant's skin pixels from sampled egocentric videos.

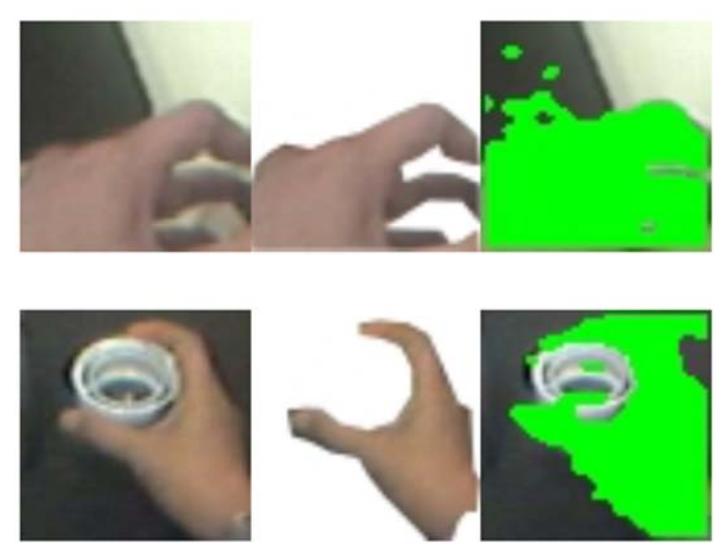

Fig. 3. Pixelwise skin classification results on selected *Gaze regions*. The *Gaze regions* (left column), Ground truths (middle column), and pixel-wise classification results (right column).

Given a pixel's UV value, $c(x, y)$, the conditional probability, $p(C_i|c)$, of $c$ belonging to a learned action $C_i$ is determined as,

$$p(C_i|c) = \frac{p(c|C_i)p(C_i)}{\sum_{j=1}^{N} p(c|C_j)p(C_j)} \tag{2}$$

Thus, the scoring function, $\Upsilon$,

$$\Upsilon = \frac{\sum_{k=1}^{56\times 56} p(C_i|C_k)}{56\times 56} \tag{3}$$

estimates the likelihood of human hand(s) to be found inside a given *Gaze region*. Two positive skin classification results are shown in Figure 3.

## V. OUR METHOD

Our method discovers temporal cuts inside a given egocentric video by tracking two motion-based parameters, $fpr$ and $sdm$ using a 9-frame median window. The median value and the 9-frame sliding window is an attempt to smooth out errors due to noise.

A simple logical AND function, $\mathrm{F}(m)$, determines a cut at frame $I_k$,

$$\mathrm{F}(I_k)\begin{cases} 1 & fpr \geq t_a \wedge sdm \geq t_b \wedge \varphi \geq t_c \wedge \Upsilon \geq t_d \\ 0 & Otherwise \end{cases} \tag{4}$$

The firstly discovered temporal cut is labeled as a *start*, the second one is labeled as an *end*. The resulting block of successive frames, encompassed by a *start* and an *end* frame, is duly registered as the first *action segment*. We repeat this process for the rest of the found temporal cuts, alternately tagging them as a *start* or an *end*.

The rest of this section describes how we calculate the two motion-based parameters: $fpr$ and $sdm$.

### *A. Motion-based Parameters*

1) Pre-processing: We start by obtaining the optical flow image for frame $I_k$. Given a Gaussian-smoothed image of the previous frame $I_{k+1}$, we determine the gradient image, $E$, in $x$, $y$ and $t$ direction,

$$\begin{aligned} E_x &= I_{k-1} \star G_x \\ E_y &= I_{k-1} \star G_y \\ E_t &= I_k - I_{k-1} \end{aligned} \tag{5}$$

where $G_x = \begin{pmatrix} +1 & 0 & -1 \\ +2 & 0 & -2 \\ +1 & 0 & -1 \end{pmatrix}$ and $G_y = \begin{pmatrix} +1 & +2 & +1 \\ 0 & 0 & 0 \\ -1 & -2 & -1 \end{pmatrix}$.

Next, we compute the velocity image for frame, $I_k$, in $u$ and $v$ direction,

$$u_k = u_{k-1} - \frac{E_x\left(E_x u_{k-1} + E_y v_{k-1} + E_t\right)}{\alpha^2 + E_x^2 + E_y^2} \tag{6}$$

$$v_k = v_{k-1} - \frac{E_y\left(E_x u_{k-1} + E_y v_{k-1} + E_t\right)}{\alpha^2 + E_x^2 + E_y^2} \tag{7}$$

where $\alpha$ is a smoothing constant.

Next, the optical flow magnitude, $\sqrt{(u_k)^2 + (v_k)^2}$, and orientation, $\tan^{-1}\frac{u_k}{v_k}$, for frame $I_k$ are estimated.

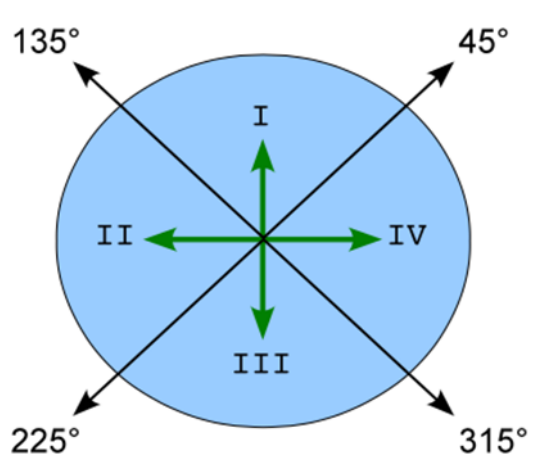

Fig. 4. The orientation bins: I, II, III and IV.

Later, we populate an optical flow histogram for frame $I_k$ with pixel-wise orientations, using the corresponding magnitude value as weightage. The four orientation bins are shown in Figure 4. The first and second most-populated bins are declared as the *background motions*. All pixels inside frame $I_k$ are then labelled as either background or foreground by comparing their optical flow orientation against the two declared *background motions*. At each frame, the total foreground pixels, $\beta_{foreground}$, and the total background pixels, $\beta_{background}$ are estimated.

2) Foreground Pixel Ratio: This parameter relates to the discontinuities between two *action segments*. *Gaze regions* belonging to a learned action are dominated by foreground pixels, $\beta_{foreground}$, i.e., pixels belonging to the active hand(s) or the manipulated object(s). For *Gaze regions* belonging to a rest state, i.e., between two separate actions, the number of foreground pixels should diminish greatly. This is in agreement with our definition of learned actions as a verb-object pair.

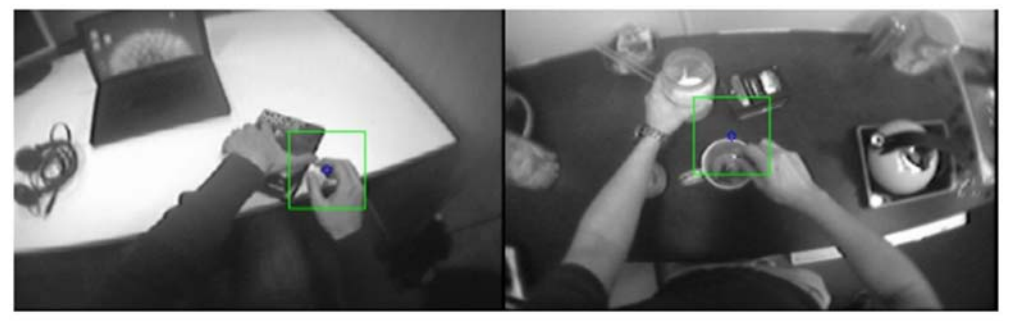

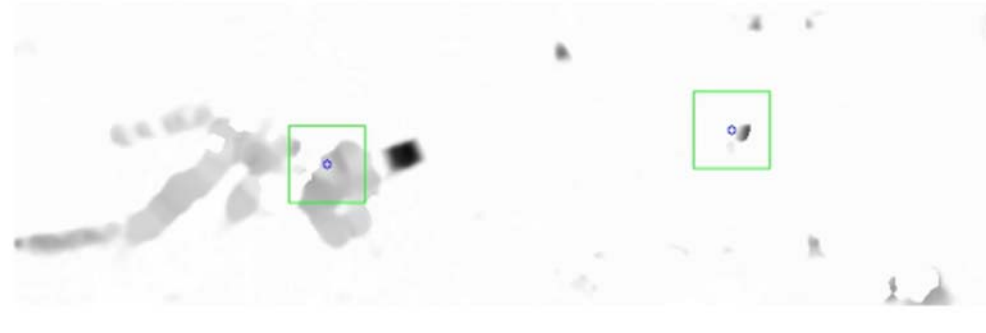

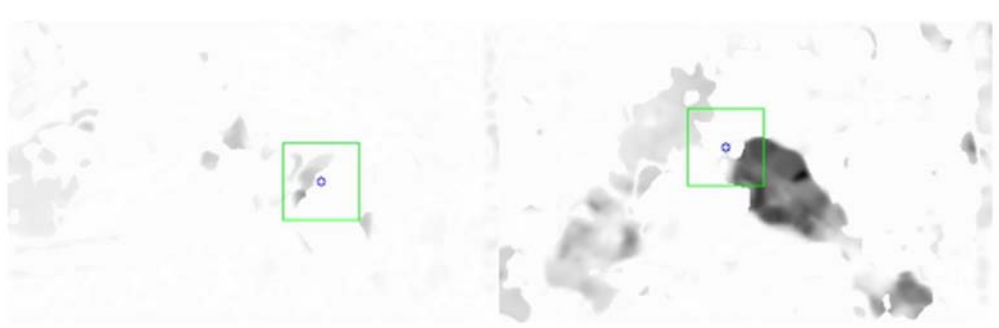

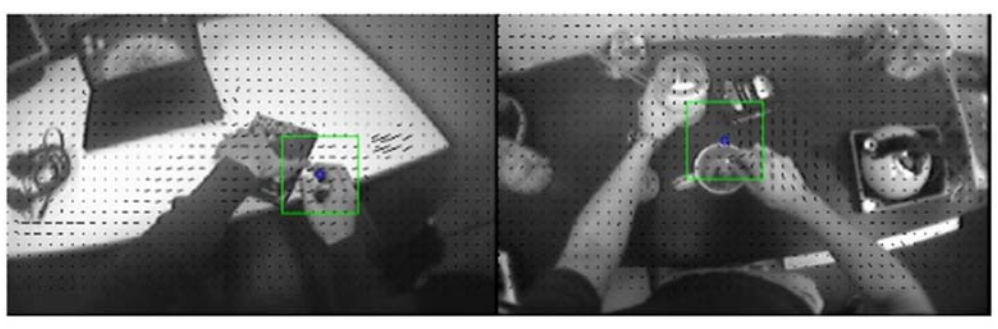

Fig. 5. Pre-processing outputs for two sample frames, each belonging to a different activity: Prepare to watch a DVD (left) and Prepare a cup of tea (right). The outputs are arranged as follows: the original frames (1st row), the horizontal magnitude images (2nd row), the vertical magnitude images (3rd row), and the visualised optical flows (4th row).

Foreground Pixel Ratio, $fpr$, is calculated as,

$$\frac{\sum \beta_{foreground}}{56 \times 56} \tag{8}$$

3) Standard Deviation of Motions: This parameter relates to the high variance of motions typically observed inside *Gaze regions* belonging to a learned action. Motions are more varied, in both orientation and magnitude, due to the presence of active hand(s) and manipulated object(s). Motions inside *Gaze regions* belonging to a rest state are typically more subdued.

Standard Deviation of Motions, $sdm$, is calculated as,

$$\frac{\sqrt{\sum_{i=0}^{4}\left(\Theta(i)-\left(\sum_{j=0}^{4}\Theta(j)/4\right)\right)^2}}{4} \tag{9}$$

where $\Theta(n)$ refers to the $n$–th orientation bin's value.

## VI. EXPERIMENTS & RESULTS

We report on a series of experiments to show the effectiveness of our proposed method. We tested several combinations of $fpr$ and $sdm$ against the ground truth data of learned actions from the BRISGAZE-ACTIONS dataset [10]. For each combination, we produced a set of temporal cuts for each egocentric video using Equation 4. The temporal cuts are then used to generate *action segments*.

To measure the segmentation accuracy, we manually obtained $n$ *action segment(s)*, $s_i^{Gr_j}$, where $i = 0, \ldots, n$, that are aligned to each ground truth segment, $Gr_j$. An *action segment* is aligned to a ground truth segment if it shares at least 50% of its frames with the ground truth segment's. Although we are ignoring the discovered action segments during rest states, we argue that these segments are easily diminishable during matching later as their underlying structure of data should be markedly different from the learned segments'.

We then measure,

$$AlignScore = \frac{\sum_{i=1}^{n} Gr_j \cap s_i^{Gr_j}}{\sum_{i=1}^{n} Gr_j \cup s_i^{Gr_j}} \qquad (10)$$

and accept the alignment as a True Positive, $TP$, if $AlignScore \geq 0.5$, or a False Positive, $FP$, otherwise. An unpaired ground truth is counted as a False Negative, $FN$. We repeat this process for all ground truth segments inside all egocentric videos.

### *A. Recall vs. Precision*

Recall, $TP/(TP + FN)$, is the fraction of retrieved ground truths and Precision, $TP/(TP + FP)$, is the fraction of retrieved ground truths that are usable, i.e., where $AlignScore \geq 0.5$.

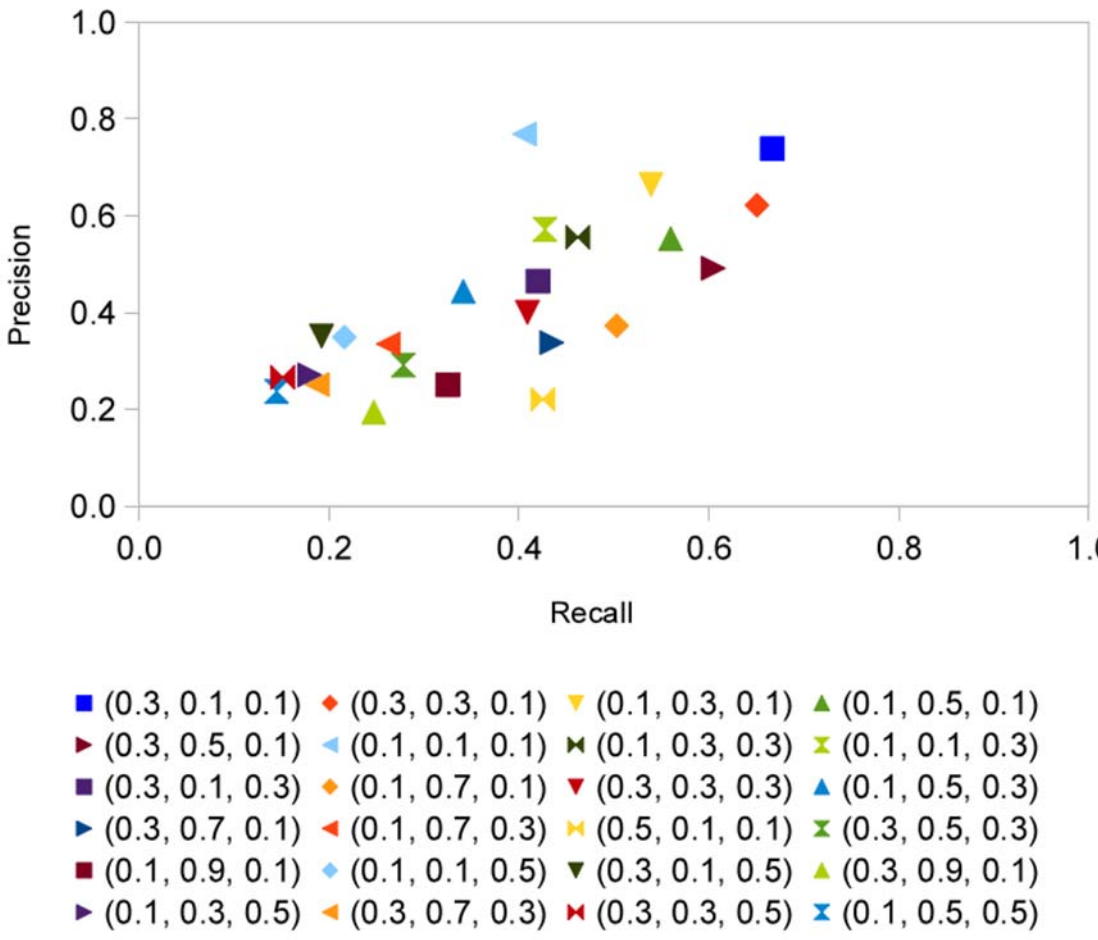

Fig. 6. Recall vs. Precision, using a fixed $\varphi$ value of 0.05. The rest of the parameters, i.e., $fpr$, $sdm$ and $\Upsilon$ are varied from 0.10 to 0.90. For visual clarity, only the top 24 combinations are shown here.

In Figure 6, the optimal combination is $fpr = 0.30$, $sdm = 0.10$, and $\Upsilon = 0.10$. Using this combination, the recall rate is 0.67, precision rate is 0.74 and an F-measure, $2 \cdot \frac{recall \cdot precision}{recall + precision}$, of 0.70.

Parameter $fpr$ requires a relatively higher threshold value of 0.30 due to the high number of misclassifications when tagging foreground pixels. This is as expected since background disambiguation of egocentric videos is a non-trivial task due the moving camera [33]. A large camera motion may introduce an out-of-plane transformation to certain background features, causing them to be erroneously tagged as foreground.

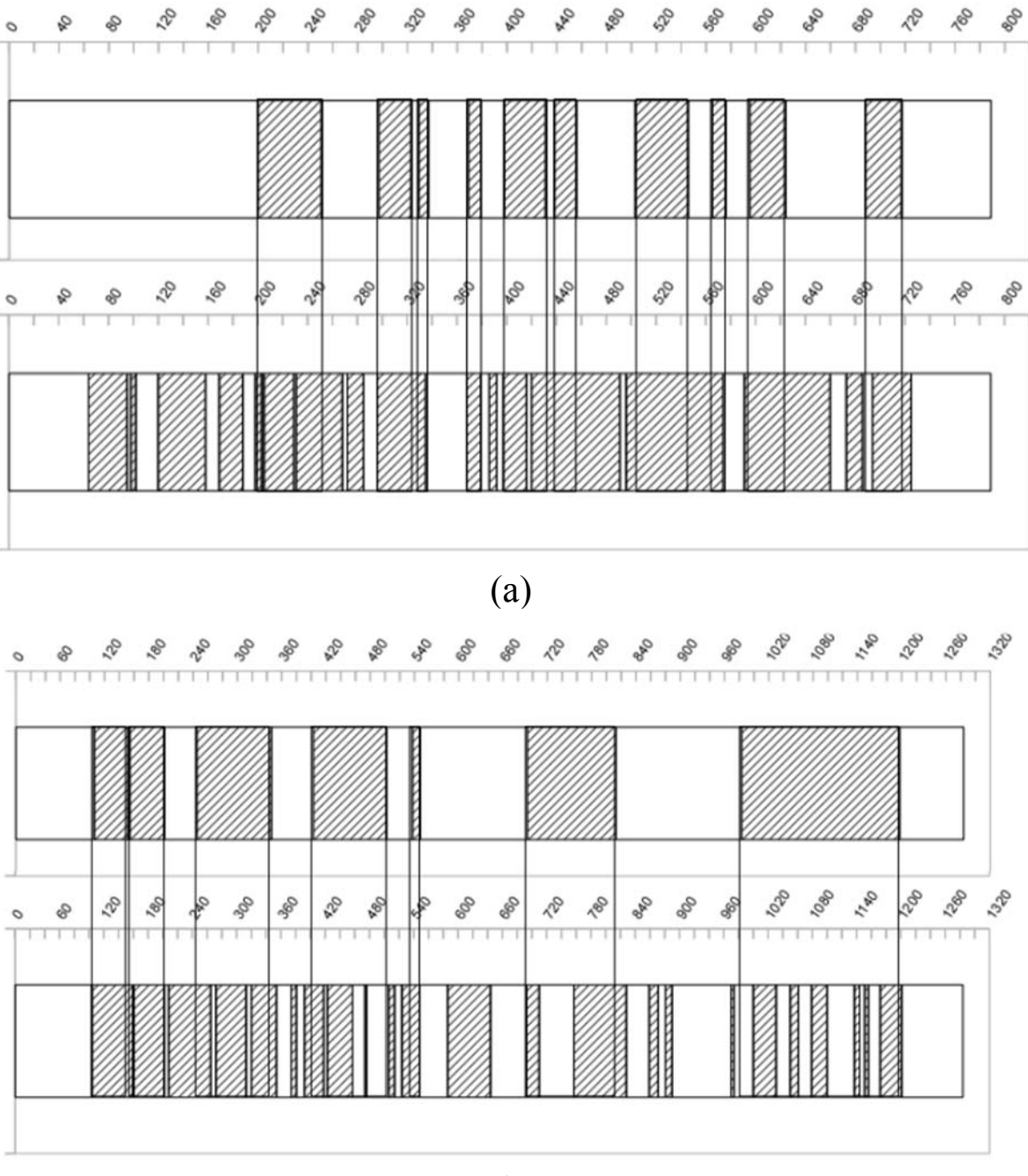

(b)

Fig. 7. Sample temporal segmentation result for activity a) "Prepare a cup of tea" and b) "Prepare to watch a DVD", both obtained using the optimal combination in Figure 6. We included both Ground truth data (top row) and *action segments* (bottom row).

Parameter $sdm$ requires a weaker threshold value of 0.10. Not all actions involve large and varied motions of the active hand(s) and the manipulated object(s). For example, action "Press key combo" in activity "Prepare to watch a DVD" only involve a small motion of the subject's finger(s) whilst the palm and the manipulated object, i.e., the keyboard, remain stationary. Consequently, the resulting *Gaze regions* exhibit a low standard deviation of motions.

Parameter $\Upsilon$ also require a weaker threshold value of 0.10. For some actions, the corresponding *Gaze regions* contain minimal skin pixels. On some cases, only a small portion of the subject's fingers are visible inside the *Gaze region*. Setting the threshold to a higher value will result to these good *Gaze regions* to be erroneously discarded.

Selected temporal segmentation results are shown in Figure 7.

## VII. CONCLUSIONS

Our method, with an average F-measure performance of 0.70, compares favorably to Kitani et al.'s, which obtained an average F-measure performance of 0.72 for their PARK sequence [16]. Poleg et al. [32] uses a different performance measure hence a direct comparison is not possible.

As evidenced by both temporal segmentation results shown in Figure 7, our proposed method managed to obtain a reasonably good segmentation output. The combination of two motion-based parameters and two entropy measures works well in discovering temporal cuts that reasonably matched the natural cuts, at the cost of over-segmentation.

The over-segmentation problem can be solved by implementing a similar coarse-to-fine representation (i.e., temporal pyramids) to [31], in which an optimal cluster containing over-segmented segments can be partially matched to a learned action. Matching of over-segmented segments can be done using a modular voting approach in [39], [40], [41] and [42], whereby a majority vote is used to determine the best-matching learned action to the optimal cluster of segments.

As for false action segments found at rest states, they are easily diminishable during matching later as the underlying structure of data should be markedly different from the learned segments'.

## ACKNOWLEDGMENT

This work was supported by the Malaysian Ministry of Education through the following RU-RACE grant: RACE/c(2)/1253/2015(09).